\documentclass{article} 
\usepackage{iclr2026_conference,times}

\usepackage{amsmath,amsfonts,bm}

\def\eqref#1{equation~\ref{#1}}

\def\1{\bm{1}}

\DeclareMathAlphabet{\mathsfit}{\encodingdefault}{\sfdefault}{m}{sl}
\SetMathAlphabet{\mathsfit}{bold}{\encodingdefault}{\sfdefault}{bx}{n}

\usepackage{hyperref}
\usepackage{url}
\usepackage{graphicx}
\usepackage{booktabs}
\usepackage{csquotes}
\usepackage{multirow}
\usepackage{makecell}
\usepackage{caption}    
\usepackage{hyperref}
\usepackage{wrapfig}
\usepackage{amsthm}
\usepackage{tabularx}
\usepackage{tikz}
\usetikzlibrary{decorations.pathreplacing}
\usepackage{xcolor}

\title{Post-Training Quantization of OpenPangu Models for Efficient Deployment on Atlas A2}

\author{
     {Yilun Luo}\textsuperscript{\rm 1}, 
     {Huaqing Zheng}\textsuperscript{\rm 1},
     {Haoqian Meng}\textsuperscript{\rm 1},
     {Wenyuan Liu}\textsuperscript{\rm 1},
     {Peng Zhang}{\textsuperscript{\rm 1}}{\thanks{Corresponding Author: Peng Zhang \textless pzhang@tju.edu.cn\textgreater}}~~\\ 
     \textsuperscript{\rm 1} School of Computer Science and Technology, Tianjin University\\
}

\iclrfinalcopy 
\begin{document}

\maketitle

\begin{abstract}
Huawei's openPangu-Embedded-1B and openPangu-Embedded-7B are variants of the openPangu large language model, designed for efficient deployment on Ascend NPUs. The 7B variant supports three distinct Chain-of-Thought (CoT) reasoning paradigms, namely slow\_think, auto\_think, and no\_think, while the 1B variant operates exclusively in the no\_think mode, which employs condensed reasoning for higher efficiency. Although CoT reasoning enhances capability, the generation of extended reasoning traces introduces substantial memory and latency overheads, posing challenges for practical deployment on Ascend NPUs. This paper addresses these computational constraints by leveraging low-bit quantization, which transforms FP16 computations into more efficient integer arithmetic. We introduce a unified low-bit inference framework, supporting INT8 (W8A8) and W4A8 quantization, specifically optimized for openPangu-Embedded models on the Atlas A2. Our comprehensive evaluation on code generation benchmarks (HumanEval and MBPP) demonstrates the efficacy of this approach. INT8 quantization consistently preserves over 90\% of the FP16 baseline accuracy and achieves a 1.5x prefill speedup on the Atlas A2. Furthermore, W4A8 quantization significantly reduces memory consumption, albeit with a moderate trade-off in accuracy. These findings collectively indicate that low-bit quantization effectively facilitates efficient CoT reasoning on Ascend NPUs, maintaining high model fidelity.
\end{abstract}

\section{Introduction}

\label{sec:introduction}
Large language models are increasingly being deployed on edge devices to facilitate on-device reasoning and code generation \citep{zheng2025reviewedgelargelanguage,Girija_2025}. Recently, Huawei introduced the openPangu-Embedded series, which includes the 7B and 1B variants \citep{chen2025panguembeddedefficientdualsystem}. The 7B variant supports three Chain-of-Thought (CoT) reasoning paradigms including slow\_think, auto\_think, and no\_think \citep{wei2023chainofthoughtpromptingelicitsreasoning,zhang2022automaticchainthoughtprompting}, which differ in the verbosity and structure of reasoning steps, thereby influencing latency and output quality. The 1B variant implements only the no\_think paradigm, producing condensed reasoning by default. Despite the strong performance of these models, their original FP16 versions require substantial memory and computational resources, which limit their practical deployment on NPU accelerators \citep{xiao2024smoothquantaccurateefficientposttraining,ashkboos2024quarot,zhao2024atomlowbitquantizationefficient}, such as the Huawei Atlas A2, where memory bandwidth and storage capacity present significant constraints.

To address the high memory and computational demands of FP16 inference, we apply quantization to the openPangu-Embedded models. This technique replaces floating-point parameters and activations with low-bit integer representations, significantly reducing both model size and computational complexity while aiming to maintain task performance \citep{wei2023outliersuppressionaccuratequantization,yao2023zeroquantv2exploringposttrainingquantization,liu2024crossquantposttrainingquantizationmethod}. Specifically, we focus on two quantization configurations commonly used for efficient large language model deployment, namely INT8 (W8A8) and W4A8  \citep{dettmers2022llmint88bitmatrixmultiplication,lee2024enhancingcomputationefficiencylarge}. Both configurations leverage the integer arithmetic capabilities of Ascend NPUs, offering a practical trade-off between efficiency and accuracy for on-device applications.

However, most existing quantization techniques are designed and optimized for GPU architectures and are not easily transferable to Ascend NPUs\citep{wang2019haqhardwareawareautomatedquantization}. Adapting quantized models to the Atlas A2 presents significant challenges due to its unique compute primitives, memory hierarchy, and integer execution model. Efficient low-bit inference on this architecture requires tight integration between the quantization format, memory layout, and compute primitives to minimize redundant data movement and avoid underutilizing hardware resources \citep{feng2022tensorirabstractionautomatictensorized,frantar2023gptqaccurateposttrainingquantization}. To facilitate practical evaluation, we develop hardware-aware quantized inference paths that are fully optimized for the Atlas A2 execution environment.

Beyond system-level deployment, we observe that quantization impacts the reasoning behavior of the openPangu-Embedded models. The three CoT paradigms slow\_think, auto\_think, and no\_think exhibit distinct characteristics and respond differently to low-bit conversion\citep{chen2025panguembeddedefficientdualsystem}. In slow\_think, the model generates detailed intermediate steps to support complex reasoning. In no\_think, the reasoning process is more condensed, retaining the stepwise structure but operating with higher efficiency. The auto\_think mode adaptively switches between these two strategies based on input characteristics. Our evaluation reveals that quantization affects the behavior of all three modes, prompting a systematic investigation into the impact of quantization on both model execution and reasoning quality.

Our work makes three practical contributions:

\textbf{(1) A unified low-bit inference framework for INT8 and W4A8 on Ascend NPUs.}
We enable efficient low-bit inference for openPangu-Embedded by implementing a hardware-aware execution framework on the Atlas A2. This framework supports both INT8 and W4A8 precisions, achieving higher throughput compared to non-optimized baselines, while eliminating performance bottlenecks caused by format conversion and memory overhead.

\textbf{(2) Assessment of mainstream quantization methods for openPangu on Ascend.}
We evaluate the accuracy of openPangu-Embedded models under several widely-used quantization schemes, including SmoothQuant and Hadamard rotation. Our assessment demonstrates that these methods can maintain competitive accuracy when applied to openPangu models targeting Ascend deployment, thereby validating their practical applicability for this model-hardware pairing.

\textbf{(3) Analysis of how quantization affects CoT reasoning.}
We analyze the impact of INT8 quantization on CoT reasoning behavior in the openPangu-Embedded-7B across the three reasoning modes and in the 1B model under its native no\_think mode, using HumanEval and MBPP as evaluation benchmarks. Our findings show that the transition from FP16 to INT8 introduces minimal disruption to the reasoning process in most scenarios, indicating that quantized models can still effectively perform structured multi-step inference.

Building on the above contributions, our experimental results demonstrate that low-bit quantization enables efficient deployment of openPangu-Embedded models on Ascend NPUs with minimal loss in model fidelity. Across HumanEval and MBPP and all three CoT reasoning paradigms, INT8 quantization consistently preserves over 90\% of FP16 accuracy while achieving up to a 1.5$\times$ inference speedup and substantial memory reduction on the Atlas A2. For more aggressive compression, W4A8 introduces moderate accuracy degradation in its baseline form, but calibration-aware techniques such as SmoothQuant and Hadamard rotation significantly improve accuracy across reasoning modes, narrowing the gap to FP16. Importantly, INT8 exhibit minimal impact on CoT reasoning behavior, indicating that structured multi-step inference can be retained under low-bit execution.

\section{Preliminary}
To enable efficient inference of large language models on resource-constrained hardware, quantization techniques are used to approximate high-precision floating-point values with low-bit integer representations. This approach reduces both memory footprint and computational cost, while aiming to preserve model accuracy \citep{dettmers2022llmint88bitmatrixmultiplication, yao2023zeroquantv2exploringposttrainingquantization, jin2024comprehensiveevaluationquantizationstrategies}.

The core operation involves mapping a real-valued tensor $X$ into a discrete set of integers through linear scaling and rounding. The transformation for quantization is defined as:

\begin{equation}
    \bar{X} = \text{round}\left( \frac{X}{s} \right),
\end{equation}

where $s$ represents the scaling factor that maps the magnitude range of $X$ to the target bit-width. The scaling factor $s$ is typically derived from the maximum absolute value observed in $X$. After quantization, the values are clamped to lie within the representable integer bounds:

\begin{equation}
    s = \frac{2 \cdot \max(|X|)}{2^n - 1}, \quad
    \bar{X} = \text{clamp}\left( \left\lfloor \frac{X}{s} \right\rceil,\, -2^{n-1},\, 2^{n-1} - 1 \right),
\end{equation}

where $n$ denotes the number of bits allocated per element. This formulation assumes that the data distribution is approximately centered around zero, a common property in transformer-based language models, and is well-suited for hardware supporting signed integer arithmetic.

The fidelity of quantization is highly dependent on the granularity of the scaling factor\citep{nagel2021whitepaperneuralnetwork}. Per-tensor quantization uses a single scale for an entire tensor, offering memory efficiency but often resulting in lower accuracy\citep{bondarenko2021understandingovercomingchallengesefficient}. Per-channel quantization assigns a separate scale to each output channel of weight matrices and is widely used due to its effectiveness \citep{lin2023awq}. For activations, per-token quantization computes one scale per token across its entire feature dimension \citep{shao2024omniquantomnidirectionallycalibratedquantization}. To balance flexibility and overhead, group-wise quantization divides each tensor into fixed-size groups, each with its own scale \citep{dettmers2023qloraefficientfinetuningquantized, zhao2024atomlowbitquantizationefficient}. These strategies are essential for maintaining accuracy under low-bit constraints.

\section{Method}
We aim to enable the efficient deployment of Huawei's openPangu-Embedded models on Ascend NPUs through quantization. Our approach focuses on two main aspects. First, we implement a unified low-bit inference framework for INT8 and W4A8 on the Atlas A2. Second, we perform a systematic evaluation of quantization accuracy across the model's three built-in CoT reasoning modes slow\_think, auto\_think, and no\_think, as well as across different quantization schemes.

\subsection{Unified Low-Bit Inference Framework for INT8 and W4A8}
We implement INT8 and W4A8 quantized inference on the Atlas A2 by configuring the weight layout, dequantization logic, and matrix multiplication within the platform's operator template library, CATLASS. This integration ensures native execution without the need for intermediate format conversions, enabling end-to-end low-bit computation specifically optimized for openPangu-Embedded models.

\subsection{Quantization Configurations and Evaluation Protocol}
We evaluate three W4A8 quantization configurations to assess the impact of common calibration-aware techniques on model accuracy. The baseline configuration applies standard quantization with calibrated scales. The two enhanced variants incorporate widely used preprocessing strategies aimed at improving quantization robustness.

The first variant adopts SmoothQuant \citep{xiao2024smoothquantaccurateefficientposttraining}, which balances the quantization difficulty between activations and weights using a diagonal scaling ${S} = \text{diag}({s})$. The linear transformation is rewritten as:  
\begin{equation}
    {Y} = ({X} {S}^{-1})({S} {W}), \quad
    s_j = \frac{\max(|{X}_j|)^\alpha}{\max(|{W}_j|)^{1 - \alpha}},
\end{equation}
with $\alpha = 0.5$ in our experiments.

\begin{figure}[ht]
\centering
\centerline{\includegraphics[width=\columnwidth]{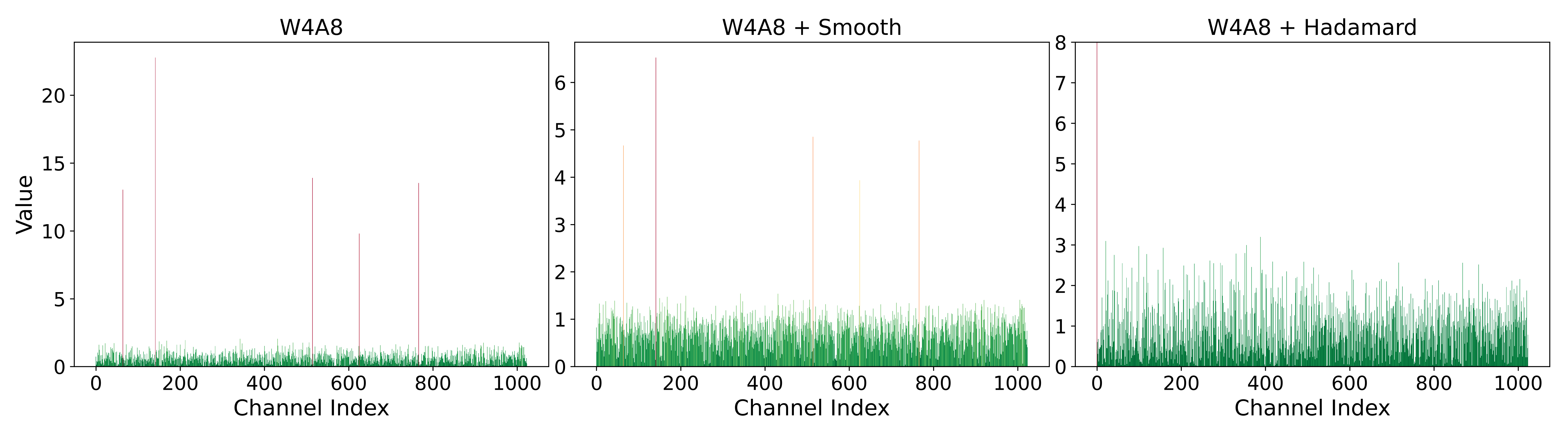}}
\caption{Channel-wise absolute value distributions under different W4A8 quantization configurations. The baseline exhibits heavy-tailed distribution with large outliers, while both SmoothQuant and Hadamard preprocessing significantly smooth the distribution.}
\label{fig:smooth_hadamard_w4a8_demo}
\end{figure}

The second variant employs Hadamard rotation \citep{ashkboos2024quarot, liu2024spinquantllmquantizationlearned}, which applies an orthogonal transform to the weight matrix before quantization. The output is computed as:
\begin{equation}
    {Y} = ({X} {H})({H}^T {W}),
\end{equation}
where ${H}$ is a normalized Hadamard matrix. This rotation preserves the mathematical equivalence in full precision while promoting a more uniform weight distribution, which is better suited for low-bit quantization. This effect is visually demonstrated in Figure~\ref{fig:smooth_hadamard_w4a8_demo}.

\section{Experiments}
\subsection{Experimental Setup}
\textbf{Quantization Configuration.}
We apply quantization without retraining, using calibrated scales derived from downstream task data. All quantized models employ symmetric quantization. Our evaluation covers INT8 and three W4A8 configurations, including a baseline W4A8 setting as well as two enhanced variants that incorporate SmoothQuant\citep{xiao2024smoothquantaccurateefficientposttraining} and Hadamard rotation\citep{ashkboos2024quarot} respectively. For performance measurement on the Atlas A2, all configurations are deployed using hardware-optimized GEMM interfaces tailored to their corresponding bitwidths. Accuracy evaluation is conducted under a unified quantization pipeline to ensure consistent calibration and quantization procedures across all settings.

\textbf{Benchmarks.}
We evaluate model performance on two standard code generation benchmarks, HumanEval \citep{human-eval} and MBPP \citep{mbpp}. For each benchmark, we test all three CoT reasoning paradigms supported by openPangu-Embedded, including slow\_think, auto\_think, and no\_think. We compare INT8 quantization against the FP16 baseline to assess accuracy retention and inference efficiency, and further analyze how the evaluated W4A8 configurations influence model fidelity under a fixed 4-bit weight setting on Ascend NPUs.

\begin{figure}[ht]
\centering
\centerline{\includegraphics[width=0.9\columnwidth]{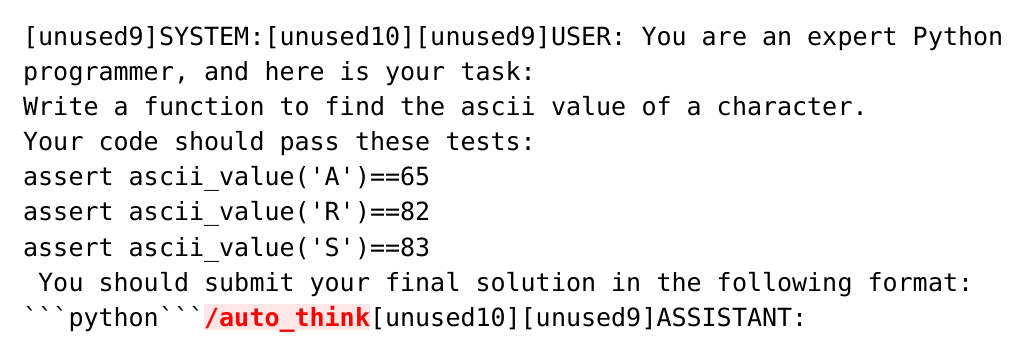}}
\caption{Example of a user prompt under the auto\_think mode. The reasoning directive no\_think (highlighted in red) is appended to the prompt to suppress explicit CoT generation.}
\label{fig:CoT_mode_switching}
\end{figure}

\textbf{CoT Mode Activation.}
For the 7B model, slow\_think is the default reasoning behavior and requires no modification to the input prompt. The auto\_think and no\_think modes are activated by appending corresponding directives (e.g., /auto\_think; see Figure~\ref{fig:CoT_mode_switching}) to the end of the input prompt. In contrast, the 1B model operates exclusively in no\_think mode by design. Its default reasoning behavior is inherently condensed and efficient, requiring no directive in the prompt.

\subsection{Main Results}
\label{subsec:main_results}
Table~\ref{tab:accuracy_comparison} reports the accuracy of openPangu-Embedded models on HumanEval and MBPP under FP16 and INT8 quantization. For the 7B model, INT8 matches or exceeds the performance of FP16 across all settings. In no\_think, the scores are similar (HumanEval: 85.37 vs 85.37; MBPP: 77.04 vs 78.21). In auto\_think, INT8 improves by 2.44 on HumanEval and 3.11 on MBPP. Under slow\_think, INT8 achieves 95.73 on HumanEval and 79.77 on MBPP, compared to FP16's 95.12 and 77.43, respectively. For the 1B model, which only supports slow\_think, INT8 slightly outperforms FP16 (HumanEval: 66.46 vs. 65.24; MBPP: 62.26 vs. 61.87). These results confirm that INT8 quantization preserves high code generation accuracy across model scales and reasoning strategies, supporting efficient deployment without significant loss in fidelity.

\begin{table}[ht]
\centering
\small
\begin{tabular}{c|c|c|cc}
\toprule
Model & CoT Mode & Precision & HumanEval & MBPP \\
\midrule
\multirow{6}{*}{7B} 
& \multirow{2}{*}{no\_think} 
& FP16 & 85.37 & 77.04 \\
& & INT8 & 85.37 & 78.21 \\
\cmidrule{2-5}
& \multirow{2}{*}{auto\_think} 
& FP16 & 92.68 & 80.16 \\
& & INT8 & 95.12 & 83.27 \\
\cmidrule{2-5}
& \multirow{2}{*}{slow\_think}
& FP16 & 95.12 & 77.43 \\
& & INT8 & 95.73 & 79.77 \\
\midrule
\multirow{2}{*}{1B}
& \multirow{2}{*}{no\_think}
& FP16 & 65.24 & 61.87 \\
& & INT8 & 66.46 & 62.26 \\
\bottomrule
\end{tabular}
\caption{openPangu-Embedded model accuracy on HumanEval and MBPP under different CoT modes and quantization schemes.}
\label{tab:accuracy_comparison}
\end{table}

Table~\ref{tab:w4a8_comparison_7b} shows the relevant results of W4A8. Due to the use of a lower 4-bit weight for weights, the accuracy of W4A8 has dropped significantly. Compared with W4A8, Smoothing and Hadamard transformation effectively improve the performance of the quantitative model by alleviating outliers.

\begin{table}[ht]
\centering
\small
\begin{tabular}{c|c|c|cc}
\toprule
Model & CoT Mode & Precision & HumanEval & MBPP \\
\midrule
\multirow{12}{*}{7B} 
& \multirow{4}{*}{no\_think} 
& FP16 & 85.37 & 77.04 \\
& & W4A8 & 81.10 & 70.04 \\
& & W4A8-smooth & 79.88 & 75.10 \\
& & W4A8-Hadamard & 80.48 & 75.49 \\
\cmidrule{2-5}
& \multirow{4}{*}{auto\_think} 
& FP16 & 92.68 &80.16 \\
& & W4A8 & 91.46 & 71.60 \\
& & W4A8-smooth & 93.90 & 72.37 \\
& & W4A8-Hadamard & 94.51 &73.15\\
\cmidrule{2-5}
& \multirow{4}{*}{slow\_think} 
& FP16 & 95.12 & 77.03 \\
& & W4A8 & 92.68 & 72.76 \\
& & W4A8-smooth & 93.29 & 73.54 \\
& & W4A8-Hadamard & 94.51 & 73.93 \\
\bottomrule
\end{tabular}
\caption{openPangu-Embedded 7B model accuracy on HumanEval and MBPP under different W4A8 configurations.}
\label{tab:w4a8_comparison_7b}
\end{table}

\subsection{Efficiency Evaluation}
\begin{table}[ht]
\centering
\small
\setlength{\tabcolsep}{4pt} 
\begin{tabularx}{\textwidth}{c|cc|cc|cc|cc|cc}
\toprule
 & \multicolumn{2}{c|}{bsz=32} & \multicolumn{2}{c|}{bsz=16} & \multicolumn{2}{c|}{bsz=8} & \multicolumn{2}{c|}{bsz=4} & \multicolumn{2}{c}{bsz=2} \\
\cmidrule(lr){2-3} \cmidrule(lr){4-5} \cmidrule(lr){6-7} \cmidrule(lr){8-9} \cmidrule(lr){10-11}
Precision & latency & memory & latency & memory & latency & memory & latency & memory & latency & memory \\
\midrule
FP16 & 5419 & 45.31 & 2852 & 30.13 & 1589 & 22.55 & 921 & 18.75 & 632 & 16.84 \\
IN8 & 3611 & 39.01 & 1975 & 23.83 & 1129 & 16.24 & 723 & 12.44 & 528 & 10.55\\
\bottomrule
\end{tabularx}
\caption{Efficiency metrics for openPangu-Embedded under INT8 vs FP16, showing prefill latency (in ms) and prefill memory (in GB) across different batch sizes.}
\label{tab:efficiency_metrics}
\end{table}

To quantify the deployment benefits of INT8 quantization on the Atlas A2, we measure prefill latency and memory consumption across batch sizes ranging from 2 to 32. As shown in Table~\ref{tab:efficiency_metrics}, INT8 consistently reduces both inference time and memory footprint compared to FP16. In terms of prefill latency, INT8 achieves up to 1.5× speedup over FP16 at a batch size of 32 and maintains a consistent advantage across all batch sizes. The speedup ratio decreases slightly as batch size reduces, falling to 1.2× at batch size = 2, reflecting the diminishing returns of low-bit computation under smaller workloads. Regarding memory usage, INT8 delivers substantial savings, ranging from 13\% to 40\% depending on the batch size. At batch size = 32, memory consumption drops from 45.31 GB to 39.01 GB, while at batch size = 2, the saving expands to 37.3\%. This trend highlights the growing benefit of quantization for memory-constrained edge deployments, where smaller batch sizes are more common. These results demonstrate that INT8 quantization not only preserves model accuracy, as shown in Section~\ref{subsec:main_results}, but also enables significant efficiency gains in real-world inference scenarios on Ascend NPUs.

\subsection{CoT Analysis}
\begin{figure}[ht]
\centering
\centerline{\includegraphics[width=\columnwidth]{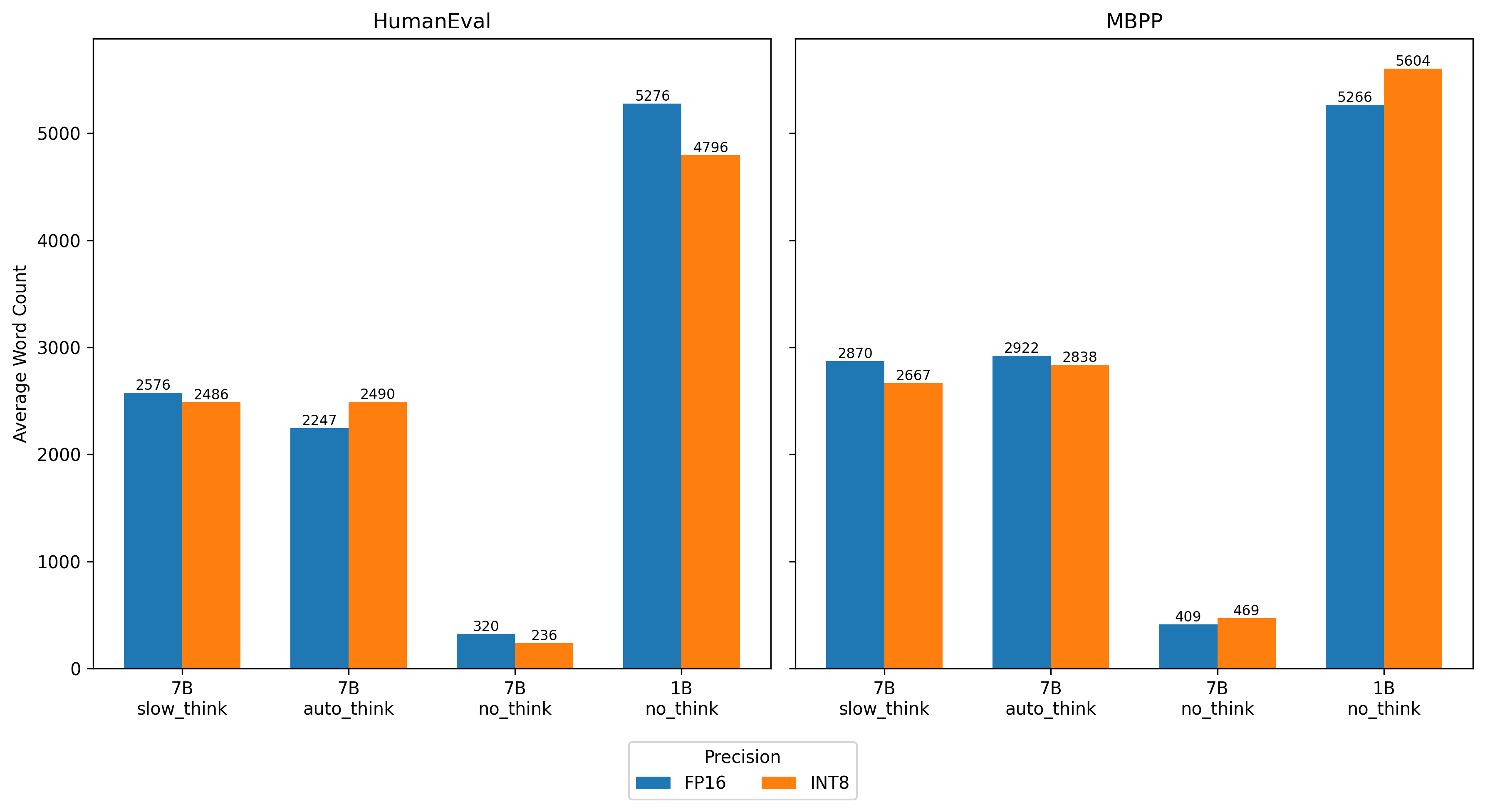}}
\caption{Average word count on HumanEval and MBPP for FP16 and INT8 openPangu-Embedded models.}
\label{fig:humaneval_mbpp_word_counts}
\end{figure}

We analyze the impact of quantization on CoT output length across the openPangu-Embedded-7B model across its three supported reasoning modes, and compare with the 1B variant operating exclusively in no\_think mode. As shown in Figure~\ref{fig:humaneval_mbpp_word_counts}, quantization from FP16 to INT8 has little effect on output length in most configurations. Notably, the 7B model consistently produces shorter CoT traces than the 1B model in no\_think mode, suggesting that larger models may adopt more concise reasoning strategies.

\begin{figure}[ht]
\centering
\centerline{\includegraphics[width=1.1\columnwidth]{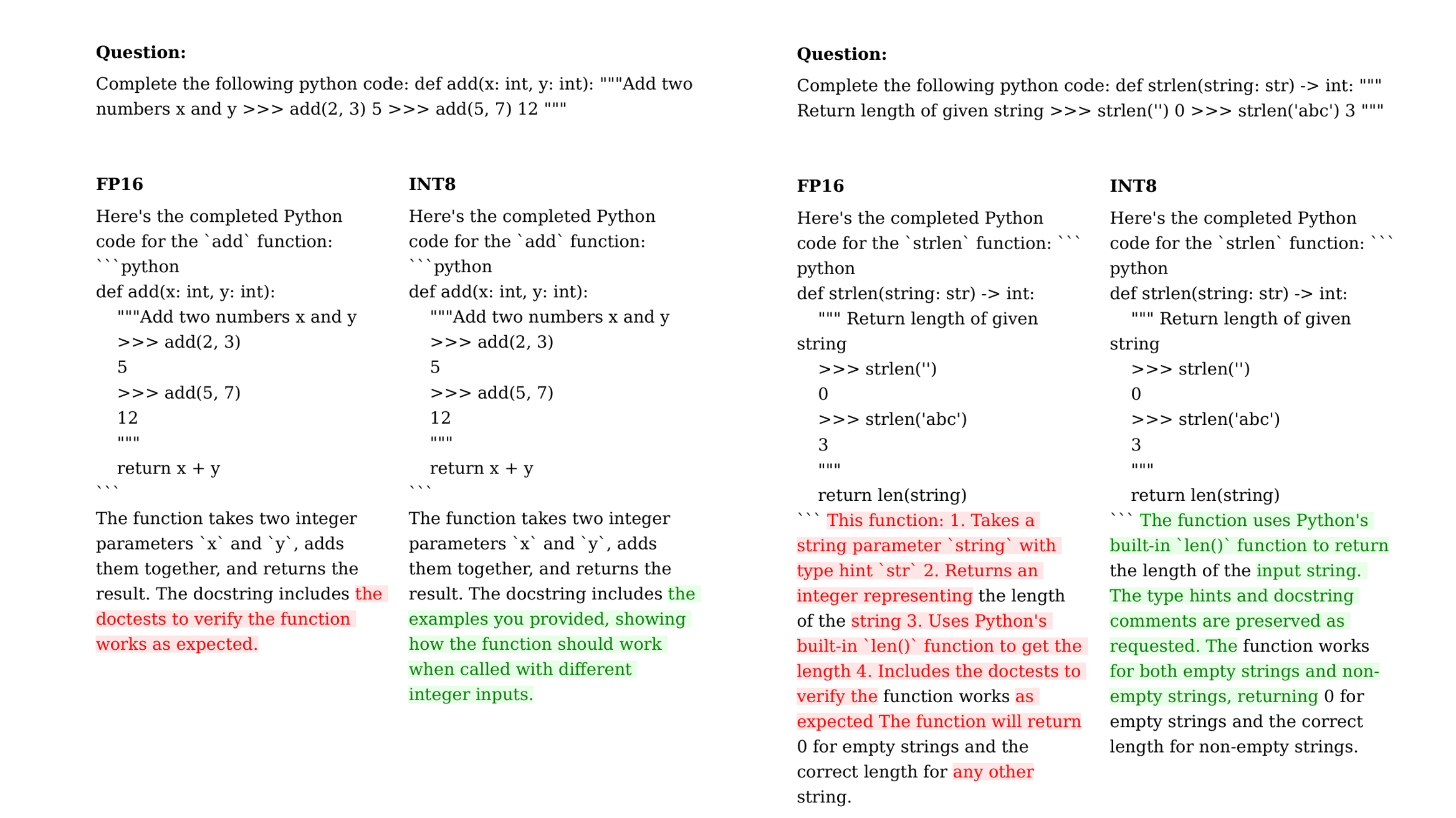}}
\caption{Qualitative comparison of CoT outputs from openPangu-Embedded-1B under FP16 and INT8 quantization. Green and red highlights indicate differences in wording or explanation depth between the two versions.}
\label{fig:CoT_demo}
\end{figure}

To complement our quantitative analysis, we examine the qualitative changes in reasoning outputs induced by quantization. Figure~\ref{fig:CoT_demo} presents side-by-side comparisons of CoT generations from the same prompts under FP16 and INT8 precision. The green and red highlights indicate differences in phrasing, explanation depth, or logical sequencing between the two versions. Notably, the core reasoning steps and final code remain functionally equivalent, suggesting that quantized models adapt their expression style to maintain performance, rather than degrade output quality.

\begin{figure}[ht]
\centering
\centerline{\includegraphics[width=0.9\columnwidth]{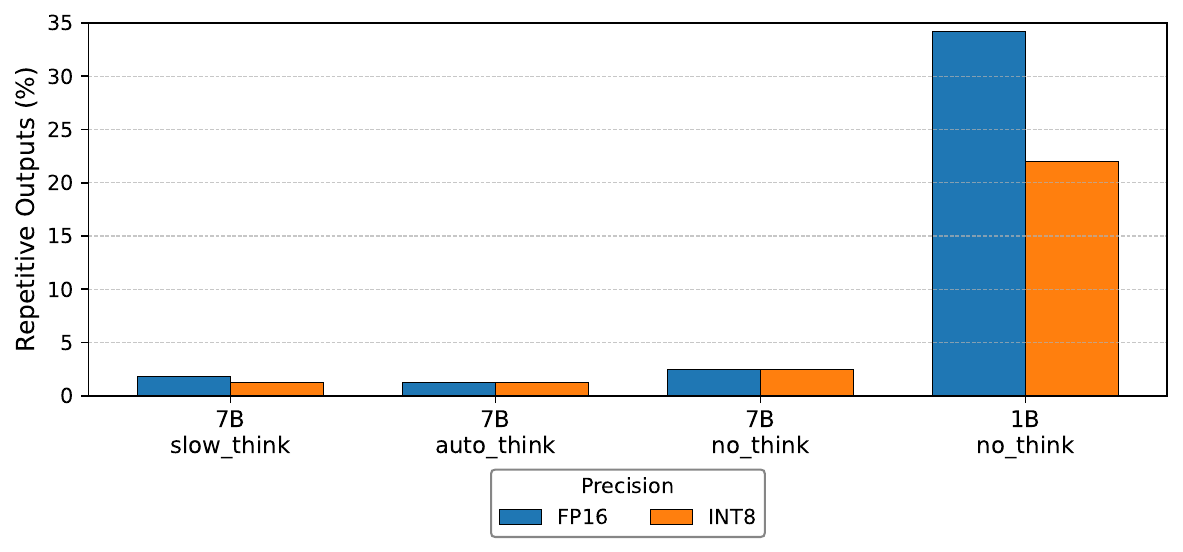}}
\caption{Frequency of repetitive generation patterns across reasoning modes and model configurations on HumanEval. Repetitive generation is defined as terminal output segments containing identical phrases repeated until sequence termination. Bars represent the percentage of test samples exhibiting this behavior under FP16 and INT8 precision for 1B (no\_think) and 7B (all three modes) model variants.}
\label{fig:repetitive_ratios}
\end{figure}

Beyond surface-level variations in output length and phrasing, we further examine the structural integrity of CoT reasoning under quantization. Specifically, we identify and analyze a critical degradation pattern termed repetitive generation, where model outputs terminate with identical phrases repeated multiple times. Figure~\ref{fig:repetitive_ratios} quantifies this phenomenon across configurations on HumanEval. The 1B model exhibits high susceptibility to repetitive outputs under FP16 (34.15\%), which is substantially reduced to 21.95\% under INT8 quantization. In contrast, the 7B model demonstrates inherent robustness across all three supported modes, with repetitive ratios consistently below 2.5\% regardless of precision or mode. Crucially, repetitive generation strongly correlates with functional failure: non-repetitive samples achieve an average accuracy of 87.39\%, while repetitive samples achieve only 18.24\%. This indicates that repetitive generation fundamentally disrupts reasoning integrity.

\section{related works}
\textbf{Post-training quantization} has become a prominent approach for compressing large language models without requiring access to original training data. It is broadly classified into weight-only quantization and weight–activation quantization. Weight-only approaches \citep{lin2023awq, frantar2023gptqaccurateposttrainingquantization, kim2024squeezellmdenseandsparsequantization} compress model weights into low-bit representations while restoring them to high precision (e.g., FP16) during GEMM computations. Although this strategy alleviates memory bandwidth demands, the underlying arithmetic operations remain in high precision, which continues to impose a substantial bottleneck on inference efficiency \citep{lin2025qservew4a8kv4quantizationcodesign}. Consequently, significant opportunities remain for accelerating large language model inference. In contrast, weight–activation methods \citep{yao2022zeroquantefficientaffordableposttraining, lee2024owqoutlierawareweightquantization} quantize both weights and activations into low-bit formats, enabling GEMM to be performed entirely in low precision. This paradigm mitigates both memory bandwidth and computational constraints but frequently incurs pronounced accuracy degradation due to outlier activations. To address this issue, mathematically equivalent transformation techniques \citep{xiao2024smoothquantaccurateefficientposttraining, shao2024omniquantomnidirectionallycalibratedquantization} employ channel-wise smoothing strategies. By transferring activation outliers into the weight domain, these methods effectively suppress quantization errors. More recently, rotation-based weight–activation methods \citep{lin2024duquant, liu2024spinquantllmquantizationlearned, ashkboos2024quarot} have been proposed and demonstrated remarkable success in maintaining model accuracy even at 4-bit precision. Complementing uniform low-bit schemes, mixed-precision quantization achieves a better accuracy-efficiency trade-off by assigning higher bitwidths to outlier-prone components and lower bitwidths to stable ones \citep{zhao2024atomlowbitquantizationefficient, liu2025micromixefficientmixedprecisionquantization}.

\textbf{Impact of quantization on reasoning behavior.} Recent studies have shown that quantization can affect the reasoning behavior of large language models beyond mere task accuracy. Empirical evidence suggests that quantization error may accumulate over CoT steps, disproportionately degrading performance on complex reasoning tasks \citep{liu2025quantizationhurtsreasoningempirical}. Furthermore, while moderate quantization typically preserves output length, which serves as a proxy for reasoning depth, aggressive low-bit schemes can lead to unstable generation and disrupted logical flow \citep{li2025quantizationmeetsreasoningexploring}. However, these analyses are primarily focused on general benchmarks and NVIDIA platforms, with limited attention given to structured reasoning modes like slow\_think or auto\_think on specialized hardware, such as Ascend NPUs.

\section{Conclusion}
In this paper, we implement a unified low-bit inference framework for INT8 and W4A8 quantization on the Huawei Atlas A2 and conduct a comprehensive study of quantization for openPangu-Embedded models on the same platform, evaluating both INT8 and W4A8 across three CoT reasoning paradigms. Our results show that INT8 quantization consistently retains over 90\% of FP16 accuracy on HumanEval and MBPP, while delivering up to 1.5× speedup and up to 40\% memory reduction. W4A8 further compresses model size, though with moderate accuracy trade-offs. Notably, both quantization schemes exhibit minimal impact on CoT reasoning, indicating preserved reasoning behavior. Together, these findings demonstrate that low-bit quantization offers practical pathways for efficient large language model deployment on Ascend-based platforms without requiring retraining. INT8 provides a balanced trade-off, while W4A8 serves as an option for extreme compression.

\bibliography{iclr2026_conference}
\bibliographystyle{iclr2026_conference}

\end{document}